\documentclass[11pt,a4paper]{article}
\usepackage[hyperref]{acl2021}
\usepackage{times}
\usepackage{soul}
\usepackage{url}
\usepackage{latexsym}
\usepackage{graphicx}
\usepackage{verbatim}
\usepackage{booktabs}
\usepackage{amssymb}
\usepackage{stfloats}
\usepackage{array}
\usepackage{bm}
\usepackage{amsfonts}
\usepackage{amsmath}
\usepackage{multirow}
\usepackage{bbding}
\usepackage{amsthm}
\usepackage{algorithm}
\usepackage{algorithmic}
\usepackage{comment}
\usepackage{color}
\usepackage{CJKutf8}
\usepackage{makecell}

\usepackage{microtype}

\aclfinalcopy 
\def\aclpaperid{1946} 

\title{Transfer Learning for Sequence Generation:\\from Single-source to Multi-source}

\author{Xuancheng Huang$^1$, Jingfang Xu$^4$, Maosong Sun$^{1,3}$, and Yang Liu${^{1,2,3}}{\thanks{\ \ Corresponding author: Yang Liu}}$
\\
$^1$Dept. of Comp. Sci. \& Tech., BNRist Center, Institute for AI, Tsinghua University \\
$^2$Institute for AI Industry Research, Tsinghua University, Beijing, China \\
$^3$Beijing Academy of Artificial Intelligence \\
$^4$Sogou Inc., Beijing, China
}

\date{}

\begin{document}
\maketitle
\begin{abstract}
\emph{Multi-source sequence generation} (MSG) is an important kind of sequence generation tasks that takes multiple sources, including automatic post-editing, multi-source translation, multi-document summarization, etc. As MSG tasks suffer from the data scarcity problem and recent pretrained models have been proven to be effective for low-resource downstream tasks, transferring pretrained sequence-to-sequence models to MSG tasks is essential.
Although directly finetuning pretrained models on MSG tasks and concatenating multiple sources into a single long sequence is regarded as a simple method to transfer pretrained models to MSG tasks, we conjecture that the direct finetuning method leads to catastrophic forgetting and solely relying on pretrained self-attention layers to capture cross-source information is not sufficient. Therefore, we propose a two-stage finetuning method to alleviate the pretrain-finetune discrepancy and introduce a novel MSG model with a fine encoder to learn better representations in MSG tasks. Experiments show that our approach achieves new state-of-the-art results on the WMT17 APE task and multi-source translation task using the WMT14 test set. When adapted to document-level translation, our framework outperforms strong baselines significantly.\footnote{The source code is available at \url{https://github.com/THUNLP-MT/TRICE}}
\end{abstract}

\section{Introduction}
Thanks to the continuous representations widely used across text, speech, and image, neural networks that accept multiple sources as input have gained increasing attention in the community \cite{ive2019distilling, dupont2000audio}. For example, multi-modal inputs that are complementary have proven to be helpful for many sequence generation tasks such as question answering \cite{antol2015vqa}, machine translation \cite{huang2016attention}, and speech recognition \cite{dupont2000audio}. In natural language processing, multiple textual inputs have also been shown to be valuable for sequence generation tasks such as multi-source translation \cite{zoph2016multi}, automatic post-editing \cite{chatterjee2017multi}, multi-document summarization \cite{haghighi2009exploring}, system combination for NMT \cite{Huang2020modeling}, and document-level machine translation \cite{wang2017exploiting}. We refer to this kind of tasks as \emph{multi-source sequence generation} (MSG).

\begin{table}[!t]
    \begin{center}
         \resizebox{\linewidth}{!}{\begin{tabular}{lcc}
            \toprule
            \bf Pretraining & \bf Single-source SG & \bf Multi-source SG \\
            \midrule
            AutoEncoding & \textsc{Bert-fused}  & \textsc{DualBert} \\
            \small (e.g., BERT) & \small \cite{zhu2019incorporating} & \small \cite{correia2019simple} \\
            \midrule
            Seq2Seq & \textsc{mBart-Trans} & \multirow{2}{*}{\em this work} \\
            \small (e.g., BART) & \small \cite{liu2020multilingual} &  \\
            \bottomrule
        \end{tabular}}
        \caption{\label{tab:intro} Comparison of various approaches to transferring pretrained models to single-source and multi-source sequence generation tasks. Different from prior studies, this work aims at transferring pretrained sequence-to-sequence models to multi-source sequence generation tasks.}
    \end{center}
\end{table}

\begin{figure*}[!t]
  \centering
  \includegraphics[width=\linewidth]{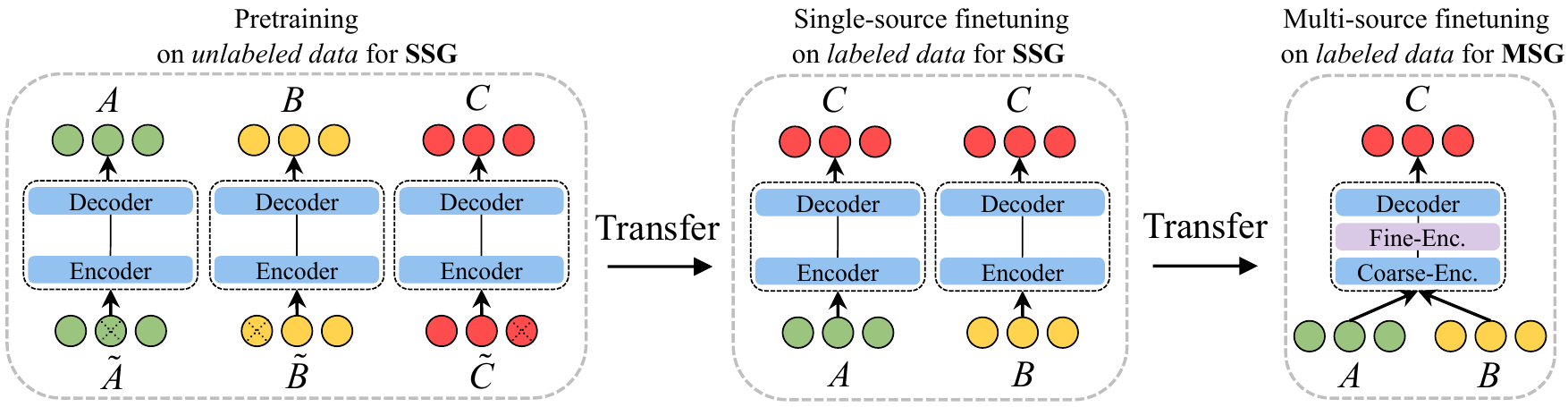}
  \caption{Overview of our framework. ``A", ``B", and ``C" denote sentences in different languages. After being pretrained on unlabeled data, the single-source sequence generation (SSG) model is finetuned on single-source labeled data. Then, the SSG model is extended to the MSG model by adding a fine encoder upon the pretrained encoder (i.e., the coarse encoder). Finally, the MSG model is finetuned on the multi-source data. The proposed framework aims to reduce the pretrain-finetune discrepancy and learn better multi-source representations.}
  \label{fig:transfer}
\end{figure*}

Unfortunately, MSG tasks face a severe challenge: there are no sufficient data to train MSG models. For example, multi-source translation requires parallel corpora involving multiple languages, which are usually restricted in quantity and coverage. Recently, as pretraining language models that take advantage of massive unlabeled data have proven to improve natural language understanding (NLU) and generation tasks substantially \cite{devlin2019bert, liu2019roberta, lewis2020bart}, a number of researchers have proposed to leverage pretrained language models to enhance MSG tasks \cite{correia2019simple, lee2020postech, lee2020cross}. For example, \citet{correia2019simple} show that pretrained autoencoding (AE) models like BERT \cite{devlin2019bert} can improve automatic post-editing.

As most recent pretrained sequence-to-sequence (Seq2Seq) models \cite{song2019mass, lewis2020bart, liu2020multilingual} have demonstrated their effectiveness in improving single-source sequence generation (SSG) tasks, we believe that pretrained Seq2Seq models can potentially bring more benefits to MSG than pretrained AE models. Although it is easy to transfer Seq2Seq models to SSG tasks, transferring them to MSG tasks is challenging because MSG takes multiple sources as the input, leading to severe \emph{pretrain-finetune discrepancies} in terms of both architectures and objectives.

A straightforward solution is to concatenate the representations of multiple sources as suggested by \citet{correia2019simple}. However, we believe this approach suffers from two major drawbacks. First, due to the discrepancy between pretraining and MSG, directly transferring pretrained models to MSG tasks might lead to catastrophic forgetting \cite{mccloskey1989catastrophic, kirkpatrick2017overcoming} that results in reduced performance. Second, the pretrained self-attention layers might not fully learn the representations of the concatenation of multiple sources because they do not make full use of the cross-source information.

Inspired by adding intermediate tasks for NLU \cite{pruksachatkun2020intermediate, vu2020exploring}, we conjecture that inserting a proper intermediate task between them can alleviate the discrepancy. In this paper, we propose a two-stage finetuning method named \textit{gradual finetuning}. Different from prior studies, our work aims to transfer pretrained Seq2Seq models to MSG (see Table \ref{tab:intro}). 
Our approach first transfers from pretrained models to SSG and then transfers from SSG to MSG (see Figure \ref{fig:transfer}). Furthermore, we propose a novel MSG model with coarse and fine encoders to differentiate sources and learn better representations. On top of a coarse encoder (i.e., the pretrained encoder), a fine encoder equipped with cross-attention layers \cite{vaswani2017attention} is added. 
We refer to our approach as TRICE (a task-agnostic Transferring fRamework for multI-sourCe sEquence generation), which achieves new state-of-the-art results on the WMT17 APE task and the multi-source translation task using the WMT14 test set. When adapted to document-level translation, our framework outperforms strong baselines significantly.

\section{Approach}
Figure \ref{fig:transfer} shows an overview of our framework. First, the problem statement is described in Section \ref{sec:statement}. Second, we propose to use the gradual finetuning method (Section \ref{sec:finetuning}) to reduce the pretrain-finetune discrepancy. Third, we introduce our MSG model, which consists of the coarse encoder (Section \ref{sec:encoder}), the fine encoder (Section \ref{sec:fine encoder}), and the decoder (Section \ref{sec:decoder}).

\subsection{Problem Statement} \label{sec:statement}
As shown in Figure \ref{fig:transfer}, there are three kinds of dataset: (1) the unlabeled multilingual dataset $\mathcal{D}_p$ containing monolingual corpora in various languages, (2) the single-source parallel dataset $\mathcal{D}_s$ involving multiple language pairs, and (3) the multi-source parallel dataset $\mathcal{D}_m$.
The general objective is to leverage these three kinds of dataset to improve multi-source sequence generation tasks. 

Formally, let $\mathbf{x}_{1:K}=\mathbf{x}_1 \dots \mathbf{x}_K$ be $K$ source sentences, where $\mathbf{x}_k$ is the $k$-th sentence. We use $x_{k,i}$ to denote the $i$-th word in the $k$-th source sentence and $\mathbf{y}=y_1 \dots y_J$ to denote the target sentence with $J$ words. The MSG model is given by
\begin{equation}
\begin{gathered}
P_m(\mathbf{y}|\mathbf{x}_{1:K}; \bm{\theta}) = \prod_{j=1}^{J}P(y_j|\mathbf{x}_{1:K},\mathbf{y}_{<j}; \bm{\theta}), \label{eqn:overall_prob}
\end{gathered}
\end{equation}
where $y_j$ is the $j$-th word in the target, $\mathbf{y}_{<j} = y_1 \dots y_{j-1}$ is a partial target sentence, $P(y_j|\mathbf{x}_{1:K},\mathbf{y}_{<j}; \bm{\theta})$ is a word-level generation probability, and $\bm{\theta}$ are the parameters of the MSG model.

\subsection{Gradual Finetuning} \label{sec:finetuning}

As training neural models on large-scale  unlabeled datasets is time-consuming, it is a common practice to utilize pretrained models to improve downstream tasks by using transfer learning methods \cite{devlin2019bert}. As a result, we focus on leveraging single-source and multi-source parallel datasets to transfer pretrained Seq2Seq models to MSG tasks. 

Curriculum learning \cite{bengio2009curriculum} aims to learn from examples organized in an easy-to-hard order, and intermediate tasks \cite{pruksachatkun2020intermediate, vu2020exploring} are introduced to alleviate the pretrain-finetune discrepancy for NLU. Inspired by these studies, we expect that changing the training objective from pretraining to MSG gradually can reduce the difficulty of transferring pretrained models to MSG tasks. Therefore, we propose a two-stage finetuning method named gradual finetuning. The transferring process is divided into two stages (see Figure \ref{fig:transfer}). In the first stage, the SSG model is transferred from denoising auto-encoding to the single-source sequence generation task, and the model architecture is kept unchanged. In the second stage, an additional fine encoder (see Section \ref{sec:fine encoder}) is introduced to transform the SSG model to the MSG model, and the MSG model is optimized on the multi-source parallel corpus. 
 
Formally, we use $\bm{\phi}_p$ to denote the parameters of the SSG model. Without loss of generality, the pretraining process can be described as follows:
\begin{alignat}{1}
\mathcal{L}_{p}(\bm{\phi}_p) &= \frac{1}{|\mathcal{D}_p|}\sum_{\mathbf{z} \in \mathcal{D}_p} \Big(-\log P_s(\mathbf{z}|\tilde{\mathbf{z}}; \bm{\phi}_p) \Big), \\
\hat{\bm{\phi}}_p &= \mathop{\mathrm{argmin}}_{\bm{\phi}_p}{\Big\{\mathcal{L}_{p}(\bm{\phi}_p)\Big\}},
\end{alignat}
where $\mathbf{z}$ is a sentence that could be in many languages, $\tilde{\mathbf{z}}$ is the corrupted sentence obtained from $\mathbf{z}$, $P_s$ is the probability modeled by the SSG model, and $\hat{\bm{\phi}}_p$ are the learned parameters. In this way, a powerful multilingual model is obtained by pretraining on the unlabeled multilingual dataset $\mathcal{D}_p$.

Then, in the first finetuning stage, let $\bm{\phi}_s$ be the parameters of the SSG model, which are initialized by $\hat{\bm{\phi}}_p$. As the single-source parallel dataset $\mathcal{D}_s$ is not always available, we can build it from the $K$-source parallel dataset $\mathcal{D}_m$. Assume ${\left \langle {\mathbf{x}_{1:K},\mathbf{y}}\right \rangle}$ is a training example in $\mathcal{D}_m$, a training example ${\left \langle {\mathbf{x},\mathbf{y}}\right \rangle}$ in $\mathcal{D}_s$ can be constructed by sampling one source from each $K$-source training example with a probability of $1/K$. The first finetuning process is given by
\begin{alignat}{1}
\mathcal{L}_{s}(\bm{\phi}_s) &= \frac{1}{|\mathcal{D}_s|}\sum_{{\left \langle {\mathbf{x},\mathbf{y}}\right \rangle \in \mathcal{D}_s}} \Big(-\log P_s(\mathbf{y}|\mathbf{x}; \bm{\phi}_s) \Big), \\
\hat{\bm{\phi}}_s &= \mathop{\mathrm{argmin}}_{\bm{\phi}_s}{\Big\{\mathcal{L}_{s}(\bm{\phi}_s)\Big\}},
\end{alignat}
where $\hat{\bm{\phi}}_s$ are the learned parameters. The learned SSG model is capable of taking inputs in multiple languages.

\begin{figure*}[!t]
  \centering
  \includegraphics[scale=0.56]{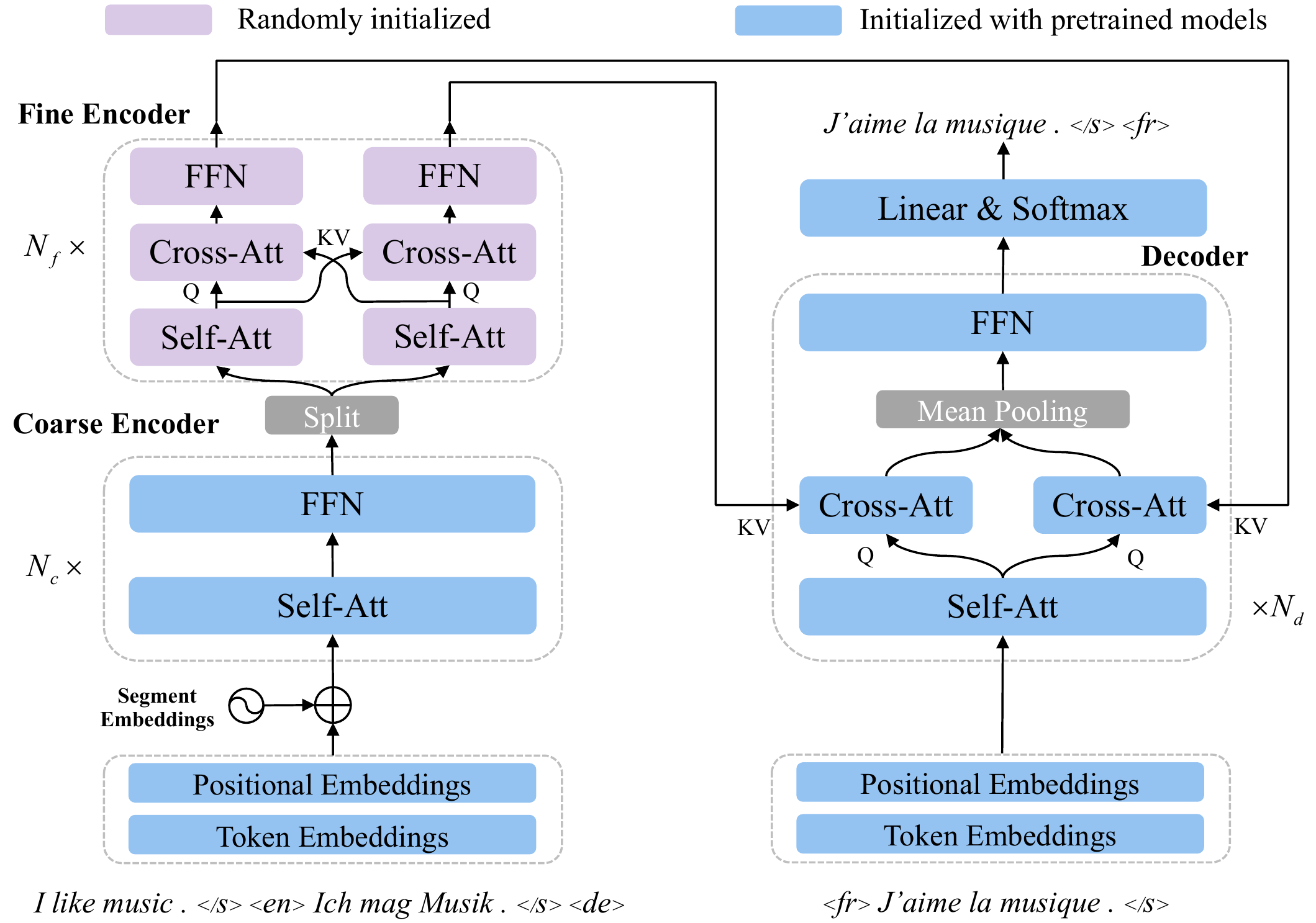} 
  \caption{The architecture of our framework. Multiple sources are first concatenated and encoded by the coarse encoder and then encoded by the fine encoder to capture fine-grained cross-source information. Finally, the representations are utilized by the decoder to generate the target sentence. For simplicity, this figure only illustrates the situation that the input contains two sources ($K=2$).}
  \label{fig:arch}
\end{figure*}

In the second finetuning stage, $\bm{\phi}_m$, the parameters of the coarse encoder, the decoder, and the embeddings, are initialized by $\hat{\bm{\phi}}_s$ while $\bm{\gamma}$ are the randomly initialized parameters of the fine encoder. Thus, $\bm{\theta}=\bm{\phi}_m \cup \bm{\gamma}$ are the parameters of the MSG model. The second finetuning process can be described as
\begin{align}
&\mathcal{L}_{m}(\bm{\theta}) \nonumber \\
= &\frac{1}{|\mathcal{D}_m|}\sum_{{\left \langle {\mathbf{x}_{1:K},\mathbf{y}}\right \rangle \in \mathcal{D}_m}} \Big(-\log P_m(\mathbf{y}|\mathbf{x}_{1:K}; \bm{\theta}) \Big), \\
& \qquad\qquad \hat{\bm{\theta}} = \mathop{\mathrm{argmin}}_{\bm{\theta}}{\Big\{\mathcal{L}_{m}(\bm{\theta})\Big\}},
\end{align}
where $P_m$ is given by Eq.~(\ref{eqn:overall_prob}). As a result, the model is expected to learn from abundant unlabeled data and perform well on the MSG task. In the following subsections, we will describe the MSG model architecture (see Figure \ref{fig:arch}) applied in the second finetuning stage.

\subsection{Input Representation and the Coarse Encoder} \label{sec:encoder}
In general, pretrained encoders are considered as strong feature extractors to learn meaningful representations \cite{zhu2019incorporating}. For this reason, \citet{correia2019simple} propose to use the pretrained multilingual encoder to encode the bilingual input pair of APE. Since MSG tasks usually have multiple sources involving different languages and pretrained multilingual Seq2Seq models like mBART \cite{liu2020multilingual} usually rely on special tokens (e.g., $<$en$>$) to differentiate languages, concatenating multiple sources into a single long sentence will make the model confused about the language of the concatenated sentence (see Table \ref{tab:ablation}). Therefore, we propose to add additional segment embedding to differentiate sentences in different languages and encode source sentences jointly by a single pretrained multilingual encoder.

Formally, the input representation can be denoted by
\begin{alignat}{1}
\mathbf{X}_{k,i} &= \mathbf{E}^{\mathrm{tok}}[x_{k,i}] + \mathbf{E}^{\mathrm{pos}}[i] + \mathbf{E}^{\mathrm{seg}}[k],
\end{alignat}
where $\mathbf{X}_{k,i}$ is the input representation of the $i$-th word in the $k$-th source sentence, and $\mathbf{E}^{\mathrm{tok}}$, $\mathbf{E}^{\mathrm{pos}}$, and $\mathbf{E}^{\mathrm{seg}}$ are the token, position, and segment/language embedding matrices, respectively. $\mathbf{E}^{\mathrm{tok}}$ and $\mathbf{E}^{\mathrm{pos}}$ are initialized by pretrained embedding matrices. $\mathbf{E}^{\mathrm{seg}}$ is implemented as constant sinusoidal embeddings \cite{vaswani2017attention}, which is denoted by $\mathbf{E}^{\mathrm{seg}}[k]_{2i}=\sin(1000*k/10000^{2i/d})$, where  $\mathbf{E}^{\mathrm{seg}}[k]_{2i+1}$ is similar to $\mathbf{E}^{\mathrm{seg}}[k]_{2i}$ and $i$ is the dimension index while $d$ is model dimension.\footnote{If the pretrained model already contains the segment/language embedding matrix, then the pretrained one is used.}

Then, the pretrained encoder is utilized to encode multiple sources:
\begin{equation}
\begin{gathered}
\mathbf{R}_{1:K}^{(i)} = \mathrm{FFN}\left(\mathrm{SelfAtt}\left(\mathbf{R}_{1:K}^{(i-1)}\right)\right), \label{eqn:encoding}
\end{gathered}
\end{equation}
where $\mathrm{SelfAtt}(\cdot)$ and $\mathrm{FFN}(\cdot)$ are the self-attention and feed-forward networks, respectively. $\mathbf{R}_{1:K}^{(i)}$ is the representation output by the $i$-th encoder layer, and $\mathbf{R}_{1:K}^{(0)}$ refers to $\mathbf{X}_{1} \dots \mathbf{X}_{K}$, where $\mathbf{X}_{k}$ is equivalent to $\mathbf{X}_{k,1} \dots \mathbf{X}_{k, I_k}$ and $I_k$ is the number of tokens in the $k$-th source sentence.

However, we conjecture that indiscriminately modeling dependencies between words by the pretrained self-attention layers cannot capture cross-source information adequately. To this end, we regard the pretrained encoder as the coarse encoder and introduce a novel fine encoder to learn better multi-source representations.

\subsection{The Fine Encoder} \label{sec:fine encoder}
To alleviate the pretrain-finetune discrepancy, we adopt the gradual finetuning method to better transfer from single-source to multi-source. In the first finetuning step, the coarse encoder is used to encode different sources individually. As multiple sources are concatenated as a single source in which words interact by pretrained self-attentions, we conjecture that the cross-source information cannot be fully captured. Hence, we propose to add a randomly initialized fine encoder, which consists of self-attentions, cross-attentions, and FFNs, on top of the pretrained coarse encoder to learn meaningful multi-source representations. Specifically, the cross-attention sublayer is an essential part of the fine encoder because they perform fine-grained interaction between sources (see Table \ref{tab:fine encoder}).

Formally, the architecture of the fine encoder can be described as follows. First, the representations of multiple sources output by the coarse encoder are divided according to the boundaries of sources:
\begin{equation}
\begin{gathered}
\mathbf{R}_{1}^{(N_c)}, \dots, \mathbf{R}_{K}^{(N_c)} = \mathrm{Split}\left(\mathbf{R}_{1:K}^{(N_c)}\right),
\end{gathered}
\end{equation}
where $N_c$ is the number of the coarse encoder layers, $\mathrm{Split}(\cdot)$ is the split operation. Second, for each fine encoder layer, the representations are fed into a self-attention sublayer:
\begin{equation}
\begin{gathered}
\mathbf{B}_{k}^{(i)} = \mathrm{SelfAtt}\left(\mathbf{A}_{k}^{(i-1)}\right),
\end{gathered}
\end{equation}
where $\mathbf{A}_{k}^{(i-1)}$ is the representation corresponding to the $k$-th source sentence output by the $(i-1)$-th layer of the fine encoder, in other words, $\mathbf{A}_{k}^{(0)}=\mathbf{R}_{k}^{(N_c)}$. $\mathbf{B}_{k}^{(i)}$ is the representation output by the self-attention sublayer of the $i$-th layer. Third, representations of source sentences interact through a cross-attention sublayer:
\begin{alignat}{1}
\mathbf{O}_{\setminus k}^{(i)} &= \mathrm{Concat}\Big(\mathbf{B}_{1}^{(i)}, \dots, \mathbf{B}_{k-1}^{(i)}, \mathbf{B}_{k+1}^{(i)}, \dots, \mathbf{B}_{K}^{(i)}\Big), \\
\mathbf{C}_{k}^{(i)} &= \mathrm{CrossAtt}\left(\mathbf{B}_{k}^{(i)}, \mathbf{O}_{\setminus k}^{(i)}, \mathbf{O}_{\setminus k}^{(i)}\right),
\end{alignat}
where $\mathrm{Concat}(\cdot)$ is the concatenation operation, $\mathbf{O}_{\setminus k}^{(i)}$ is the concatenated representation except $\mathbf{B}_{k}^{(i)}$,  $\mathrm{CrossAtt}(Q,K,V)$ is the cross-attention sublayer, $\mathbf{C}_{k}^{(i)}$ is the representation output by the cross-attention sublayer of the $i$-th layer. Finally, the last sublayer is a feedforward network:
\begin{equation}
\begin{gathered}
\mathbf{A}_{k}^{(i)} = \mathrm{FFN}\left(\mathbf{C}_{k}^{(i)}\right).  \label{eqn:output_fine encoder}
\end{gathered}
\end{equation}
After the $N_f$-layer fine encoder, the representations corresponding to multiple sources are given to the decoder.

\subsection{The Decoder} \label{sec:decoder}

Given that representations of multiple sources are different from that of a single source, to better leverage representations of multiple sources, we let the cross-attention sublayer take each source's representation as key/value separately and then combine the outputs by mean pooling.\footnote{There is little difference between the “parallel attention combination strategy” proposed by \citet{libovicky2018input} and our method.} Formally, the differences between our decoder and the traditional Transformer decoder are described below. 

First, the input representations of the $i$-th decoder layer are fed into the self-attention sublayer to obtain $\mathbf{G}_{j}^{(i)}$. Second, a \textit{separated} cross-attention sublayer is adopted by our framework to replace the traditional cross-attention sublayer:
\begin{alignat}{1}
\mathbf{P}_{j, k}^{(i)} &= \mathrm{CrossAtt}\left(\mathbf{G}_{j}^{(i)}, \mathbf{A}_{k}^{(N_f)}, \mathbf{A}_{k}^{(N_f)}\right), \\
\mathbf{H}_{j}^{(i)} &= \mathrm{MeanPooling}\left(\mathbf{P}_{j, 1}^{(i)}, \dots, \mathbf{P}_{j, K}^{(i)}\right),
\end{alignat}
where $\mathbf{A}_{k}^{(N_f)}$ is the output of the fine encoder derived by Eq.~(\ref{eqn:output_fine encoder}), $\mathbf{P}_{j, k}^{(i)}$ is the representation corresponding to the $k$-th source, $\mathbf{H}_{j}^{(i)}$ is the combined result of the separated cross-attention sublayer, and the parameters of separated cross-attentions to leverage each source are shared. Finally, a feedforward network is the last sublayer of a decoder layer. In this way, the decoder in our framework can better handle representations of multiple sources.

\begin{table*}[!t]
    \begin{center}
        \resizebox{\textwidth}{!}{
        \begin{tabular}{lcllll}
            \toprule
            \multirow{2}{*}{\bf Models} & \multirow{2}{*}{\bf Pretraining} & \multicolumn{2}{c}{\bf TEST16} & \multicolumn{2}{c}{\bf TEST17} \\
            \cmidrule(lr){3-4} \cmidrule(lr){5-6}
            & & \em \ TER & \em BLEU & \em \ TER & \em BLEU  \\
            \midrule
            \multicolumn{6}{c}{\em extremely low-resource} \\
            \midrule
            \textsc{ForcedAtt} {\small \cite{berard2017lig}} & --- & 22.89 & \ \ --- & 23.08 & 65.57 \\
            \midrule
            \textsc{DualBert} {\small \cite{correia2019simple}} & \small \em mBERT & 18.88 & 71.61 & 19.03 & 70.66 \\
            \textsc{DualBart} {\small \cite{correia2019simple}} & \small \em mBART & 18.26 & 72.65 & 18.41 & 72.08 \\
            \midrule
            \textsc{TRICE} & \small \em mBART & \textbf{17.41}$^{\vartriangle\star}$ & \textbf{73.43}$^{\vartriangle\star}$ & \textbf{17.75}$^{\vartriangle\star}$ & \textbf{72.70}$^{\vartriangle\star}$ \\
            \midrule
            \multicolumn{6}{c}{\em high-resource} \\
            \midrule
            \textsc{DualTrans} {\small \cite{junczys2018ms}} & --- & 17.81 & 72.79 & 18.10 & 71.72\\
            \textsc{L2Copy} {\small \cite{huang2019learning}} & --- & 17.45 & 73.51 & 17.77 & 72.98 \\
            \midrule
            \textsc{DualBert} {\small \cite{correia2019simple}} & \small \em mBERT & 16.91 & 74.29 & 17.26 & 73.42 \\
            \textsc{DualBart} {\small \cite{correia2019simple}} & \small \em mBART & 16.40 & 74.74 & 17.26 & 73.56 \\
            \midrule
            \textsc{TRICE} & \small \em mBART & \textbf{16.09}$^{\vartriangle\star}$ & \textbf{75.39}$^{\vartriangle\star}$ & \textbf{16.91}$^{\vartriangle\star}$ & \textbf{74.09}$^{\vartriangle\star}$ \\
            \bottomrule
        \end{tabular}}
        \caption{\label{tab:ape} Results on the automatic post-editing task (\emph{extremely low}- and \emph{high}-resource). ``\textsc{DualBart}": a method to leverage pretrained Seq2Seq models adapted from ``\textsc{DualBert}". Please refer to Appendix \ref{app:baselines} for detailed descriptions of baselines and the same below. ``$\vartriangle$": significantly better than ``\textsc{DualBert}" ($p < 0.01$). ``${\star}$": significantly better than ``\textsc{DualBart}" ($p < 0.01$).}
    \end{center}
\end{table*}

\begin{table*}[!t]
    \begin{center}
        \scalebox{1.0}{
        \begin{tabular}{lccc}
            \toprule
            \bf Models & \bf Source & \bf Pretraining & \bf TEST14 \\
            \midrule
            \textsc{MultiRnn} {\small \cite{zoph2016multi}} & \small \em (De, Fr) & --- & 30.0 \\
            \textsc{DualTrans} {\small \cite{junczys2018ms}} & \small \em (De, Fr) & --- & 37.0 \\
            \midrule
            \textsc{mBart-Trans} {\small \cite{liu2020multilingual}} & \small \em De & \small \em mBART & 31.8 \\
            \textsc{mBart-Trans} {\small \cite{liu2020multilingual}} & \small \em Fr & \small \em mBART & 34.8 \\
            \textsc{DualBart} {\small \cite{correia2019simple}} & \small \em (De, Fr) & \small \em mBART & 40.2 \\
            \midrule
            \textsc{TRICE} & \small \em (De, Fr) & \small \em mBART & \ \ \textbf{41.5}$^{\star}$ \\
            \bottomrule
        \end{tabular}}
        \caption{\label{tab:multi-source_translation} Results on the multi-source translation task ({\em medium}-resource). In this task, German and French sources are translated to English target. ``\textsc{mBart-Trans}": a single-source model directly finetuned from mBART. ``${\star}$": significantly better than ``\textsc{DualBart}" ($p < 0.01$).} 
    \end{center}
\end{table*}

\begin{table*}[!t]
    \begin{center}
        \resizebox{\linewidth}{!}{
        \begin{tabular}{lcccccc}
            \toprule
            \multirow{2}{*}{\bf Models} & 
            \multirow{2}{*}{\bf \#Context} &
            \multirow{2}{*}{\bf Pretraining} &
            \multicolumn{2}{c}{\bf TED} &
            \multicolumn{2}{c}{\bf News}\\
            \cmidrule(lr){4-5} \cmidrule(lr){6-7}
            & & & \em s-BLEU & \em d-BLEU & \em s-BLEU & \em d-BLEU \\
            \midrule
            SAN {\small \cite{maruf2019selective}} & --- & --- & 24.4 & --- & 24.8 & --- \\
            QCN {\small \cite{yang2019enhancing}} & --- & --- & 25.2 & --- & 22.4 & --- \\
            MCN {\small \cite{zheng2020towards}} & --- & --- & 25.1 & 29.1 & 24.9 & 27.0 \\
            \midrule
            \textsc{mBart-Trans} {\small \cite{liu2020multilingual}} & 0 & \small \em mBART & 28.1 & 31.7 & 29.4 & 31.2 \\
            \textsc{mBart-DocTrans} {\small \cite{liu2020multilingual}} & 1 & \small \em mBART & 28.1 & 31.7 & 28.6 & 30.2 \\
            \textsc{DualBart} {\small \cite{correia2019simple}} & 1 & \small \em mBART & 27.8 & 31.4 & 29.3 & 31.3 \\
            \midrule
            \textsc{TRICE} & 1 & \small \em mBART & \bf ~~~~\,28.5$^{\dag\ddag\star}$ & \bf ~~~~\,32.1$^{\dag\ddag\star}$ & \bf ~~~~\,29.8$^{\dag\ddag\star}$ & \bf ~~~~\,31.7$^{\dag\ddag\star}$ \\
            \bottomrule
        \end{tabular}}
        \caption{\label{tab:doc-level_MT} Results on the document-level translation task ({\em low}-resource). ``s-BLEU" and ``d-BLEU" denote BLEU scores calculated at sentence- and document-level, respectively. ``\#Context" denotes the number of context used by context-aware models. ``$\dag$": significantly better than ``\textsc{mBart-Trans}" ($p < 0.01$). ``$\ddag$": significantly better than ``\textsc{mBart-DocTrans}" ($p < 0.01$). ``$\star$": significantly better than ``\textsc{DualBart}" ($p < 0.01$).}
    \end{center}
\end{table*}

\section{Experiments}
\subsection{Setup}
\subsubsection*{Datasets}
We evaluated our framework on three MSG tasks: (1) automatic post-editing (APE), (2) multi-source translation, and (3) document-level translation.

For the APE task, following \citet{correia2019simple}, we used the data from the WMT17 APE task (English-German SMT) \cite{chatterjee2019findings}. The dataset contains 23K dual-source examples (e.g., $\langle$English source sentence, German translation, German post-edit$\rangle$) for training in an {\em extremely low}-resource setting. We also followed \citet{ correia2019simple} to adopt pseudo data \cite{junczys2016log, negri2018escape}, which contains about 8M pseudo training examples, to evaluate our framework in a {\em high}-resource setting. We adopted the {\em dev16} for development and used {\em test16} and {\em test17} for testing.

For the multi-source translation task, following \citet{zoph2016multi}, we used a subset of the WMT14 news dataset \cite{bojar2014proceedings}, which contains 2.4M dual-source examples (e.g., $\langle$German source sentence, French source sentence, English translation$\rangle$) for training, 3,000 from {\em test13} for development, and 1,503 from {\em test14} for testing.\footnote{A dual-source example can be obtained by matching two single-source examples.} It can be seen as a {\em medium}-resource setting.

For the document-level translation task, we used the dataset provided by \citet{maruf2019selective} from IWSLT2017 (TED) and News Commentary (News), both including about 200K English-German training examples, which can be seen as {\em low}-resource settings. For IWSLT2017, \emph{test16} and \emph{test17} were combined as the test set, and the rest served as the development set. For News Commentary, \emph{test15} and \emph{test16} in WMT16 were used for development and testing, respectively. We took the nearest preceding sentence as the context, and then constructed the dual-source example like $\langle$German context, German current sentence, English translation$\rangle$. 

\subsubsection*{Hyper-parameters}
We adopted mBART \cite{liu2020multilingual} as the pretrained Seq2Seq model. We set both $N_c$ and $N_d$ to 12, and $N_f$ to 1. The model dimension, the filter size, and the number of heads are the same as mBART. We adopted the vocabulary of mBART, which contains 250K tokens. We used minibatch sizes of 256, 1,024, 4,096, and 16,384 tokens for {\em extremely low-, low-, medium-}, and {\em high-}resource settings, respectively. We used the development set to tune the hyper-parameters and select the best model. In inference, the beamsize was set to 4. Please refer to Appendix \ref{app:hyper-para} for more details.

\subsubsection*{Evaluation Metrics}
We used case-sensitive BLEU ({\em multi-bleu.perl}) and TER for automatic post-editing. For multi-source translation and document-level translation, \textsc{SacreBLEU}\footnote{The signature is ``BLEU+case.mixed+numrefs.1+smooth .exp+tok.13a+version.1.4.14".} \cite{post2018call} and METEOR\footnote{\url{https://www.cs.cmu.edu/~alavie/METEOR/}} was adopted for evaluation. We used the paired bootstrap resampling \cite{koehn2004statistical} for statistical significance tests.

\begin{table}[!t]
    \begin{center}
        \scalebox{0.90}{
        \begin{tabular}{lrc}
            \toprule
            \bf Variants & \bf \#Para. & \bf BLEU \\
            \midrule
            \em None & 0M & 73.65 \\
            FFN adapter {\small \cite{guo2020incorporating}} & 100.7M & 73.71 \\
            \midrule
            Fine encoder ($N_f=1$) w/o CA & 12.5M & 73.84 \\
            Fine encoder ($N_f=1$) & 16.8M &\textbf{74.21} \\
            Fine encoder ($N_f=2$) & 33.6M & 73.70 \\
            Fine encoder ($N_f=3$) & 50.4M & 58.84 \\
            \bottomrule
        \end{tabular}}
        \caption{\label{tab:fine encoder} Comparisons with the variants of the fine encoder. ``\#Para." denotes the number of parameters and ``CA" denotes the cross-attention sublayer. ``$N_f$" denotes the number of the fine encoder layers.}
    \end{center}
\end{table}

\subsection{Main Results} 

Table \ref{tab:ape} shows the results on the automatic post-editing task. Our framework outperforms previous methods without pretraining (i.e., \textsc{ForcedAtt}, \textsc{DualTrans}, and \textsc{L2Copy}) by a large margin and surpasses strong baselines with pretraining (i.e., \textsc{DualBert} and \textsc{DualBart}), which concatenate multiple sources into a single source, significantly in both \emph{extremely low}- and \emph{high}-resource settings. Notably, the performances of our framework in the \emph{extremely low}-resource setting are comparable to results of strong baselines without pretraining in the \emph{high}-resource setting and we achieve new state-of-the-art results on this benchmark.

Table \ref{tab:multi-source_translation} demonstrates the results on the multi-source translation task. Our framework substantially outperforms both baselines without pretraining (i.e., \textsc{MultiRnn} and \textsc{DualTrans}) and with pretraining (i.e., single-source model \textsc{mBart-Trans} and dual-source model \textsc{DualBart}). Surprisingly, the single-source models with pretraining are inferior to the multi-source model without pretraining, which indicates that multiple sources play an important role in the translation task.

Table \ref{tab:doc-level_MT} shows the results on the document-level translation task. Our framework achieves significant improvements over all strong baselines. Unusually, the previous method for handling multiple sources (i.e., \textsc{DualBart}) fails to consistently outperform simple sentence- and document-level Transformer (i.e., \textsc{mBart-Trans} and \textsc{mBart-DocTrans}) while our framework outperforms these strong baselines significantly.

In general, our framework shows a strong generalizability across three different MSG tasks and four different data scales, which indicates that it is useful to alleviate the pretrain-finetune discrepancy by gradual finetuning and learn multi-source representations by fully capturing cross-source information.

\begin{table}[!t]
    \begin{center}
        \scalebox{0.96}{
        \begin{tabular}{lc}
            \toprule
            \bf Model Variants & \bf BLEU \\
            \midrule
            TRICE & \textbf{74.21} \\
            \ \ -- gradual finetuning & 73.83 \\
            \ \ -- separated cross-attention & 73.81 \\
            \ \ -- concatenated encoding & 73.61 \\
            \ \ -- segment embedding & 72.92 \\
            \bottomrule
        \end{tabular}}
        \caption{\label{tab:ablation} Ablation study. The case-sensitive BLEU scores are calculated on the development set of the APE task for all experiments for analyses. Note that we remove only one component at a time.}
    \end{center}
\end{table}

\subsection{Analyses}

In this subsection, we further conduct studies regarding the variants of the fine encoder, ablations of the other proposed components, and effect of freezing parameters. Experiments are conducted on the APE task in the \emph{extremely low}-resource setting. The BLEU scores calculated on the development set are adopted as the evaluation metric.

\paragraph{Comparisons with the variants of the fine encoder.} \label{sec:abla_fine encoders}
Table \ref{tab:fine encoder} demonstrates comparisons with the variants of the fine encoder. We find that the fine encoder (see Section \ref{sec:fine encoder}) is effective (compared to ``\emph{None}"), the cross-attention sublayer is important (compared to the one without cross-attention), and our approach outperforms ``FFN adapter", which is proposed by \citet{zhu2019incorporating} to incorporate BERT into sequence generation tasks by inserting FFNs into each encoder layer. We find that stacking more fine encoder layers even harms the performance (see the last three rows in Table \ref{tab:fine encoder})  which rules out the option that the improvements owe to increasing of parameters.

\paragraph{Ablations on the other proposed components.}  \label{sec:ablation}
Table \ref{tab:ablation} shows the results of the ablation study. We find that gradual finetuning method (see Section \ref{sec:finetuning}) is significantly beneficial. Lines ``- segment embedding" and ``- concatenated encoding" show that concatenating multiple sources into a long sequence and adding sinusoidal segment embedding for the coarse encoder are helpful (see Section \ref{sec:encoder}). The line ``- separated cross-attention" reveals that taking each source’s representation as key/value separately and then combine the outputs is better than concatenating all the representations and do the cross-attention jointly (see Section \ref{sec:decoder}).

\paragraph{Effect of freezing pretrained parameters.}  \label{sec:freezing}
As shown in Table \ref{tab:freezing}, finetuning all parameters including parameters initialized by pretrained models and parameters initialized randomly is essential for achieving good performance on MSG tasks.

\begin{table}[!t]
    \begin{center}
         \scalebox{0.96}{
        \begin{tabular}{lc}
            \toprule
            \bf Components to Finetune & \bf BLEU \\
            \midrule
            All & \textbf{74.21} \\
            The fine encoder & 70.20 \\
            \bottomrule
        \end{tabular}}
        \caption{\label{tab:freezing} Effect of freezing pretrained parameters.}
    \end{center}
\end{table}

\subsection{Adversarial Evaluation}

We adopt adversarial evaluation similar to \citet{libovicky2018input} which replaces one source with a randomly selected sentence. As shown in Table \ref{tab:adv}, both sources play important parts and the French side is more important than the German side (Randomized Fr vs. Randomized De).

\begin{table*}[!t]
    \begin{center}
         \scalebox{0.96}{
        \begin{tabular}{lcccccc}
            \toprule
            \multirow{2}{*}{\bf Models} & \multicolumn{2}{c}{\bf Normal} & \multicolumn{2}{c}{\bf Randomized Fr} & \multicolumn{2}{c}{\bf Randomized De}\\
            \cmidrule(lr){2-3} \cmidrule(lr){4-5} \cmidrule(lr){6-7}
            & \em \ BLEU & \em METEOR & \em \ BLEU & \em METEOR & \em \ BLEU & \em METEOR \\
            \midrule
            \textsc{mBart-Trans} (De) & 31.8 & 33.9 & --- & --- & --- & --- \\
            \textsc{mBart-Trans} (Fr) & 34.8 & 37.9 & --- & --- & --- & --- \\
            \textsc{DualBart} & 40.2 & 38.9 & 11.3 & 13.1 & \bf 24.9 & \bf 26.4 \\
            TRICE & \bf 41.5 & \bf 39.8 & \bf 13.5 & \bf 15.0 & 23.0 & 23.9 \\
            \bottomrule
        \end{tabular}}
        \caption{\label{tab:adv} Adversarial evaluation on the multi-source translation task. ``Randomized Fr/De" denotes that the Fr/De source is replaced with a randomly selected sentence.}
    \end{center}
\end{table*}

\subsection{Case Study}
An example in multi-source translation task is shown in Table \ref{tab:example}. The four outputs at the bottom of the table are generated by the last four models in Table \ref{tab:multi-source_translation}. We find that single-source models have different errors (e.g., ``each hospitals" and ``travelling clinics") and multi-source models fix some errors because of taking two sources. Additionally, DualBart still output erroneous ``weekly", while TRICE outputs ``weekend" successfully. We believe TRICE is better than baselines because multiple sources are complementary and the fine encoder could capture finer cross-source information, which helps correct translation errors.

\begin{table*}[!t]
    \begin{center}
        \begin{tabular}{ll}
            \toprule
            Input-Fr & Dans cet hôpital itinérant, divers soins de santé sont prodigués. \\
            Input-De & \makecell[l]{Jede dieser Wochenendkliniken bietet medizinische Versorgung in einer\\ Reihe von Bereichen an.} \\
            \midrule
            Reference-En & Each of these weekend clinics provides a variety of medical care. \\
            \midrule
            \textsc{mBart-Trans} (De) & \underline{Each weekend hospitals} offers medical care in a number of areas. \\
            \textsc{mBart-Trans} (Fr) & \underline{This travelling clinics} provides a variety of healthcare services. \\
            \textsc{DualBart} & Each of these \underline{weekly} hospitals provides healthcare in a variety of areas. \\
            TRICE & Each of these weekend clinics offers a variety of health care. \\
            \bottomrule
        \end{tabular}
        \caption{\label{tab:example} Example of multi-source translation. Some \underline{erroneous parts} are highlighted by underlines. \textsc{mBart-Trans} (De/Fr) takes single source (De/Fr) as input while \textsc{DualBart} and TRICE take both sources as input. We believe that multiple sources are complementary and TRICE could correct errors by capturing finer cross-source information.}
    \end{center}
\end{table*}

\section{Related Work}

\subsection{Multi-source Sequence Generation} 
Multi-source sequence generation includes multi-source translation \cite{zoph2016multi}, automatic post-editing \cite{chatterjee2017multi}, multi-document summarization \cite{haghighi2009exploring}, system combination for NMT \cite{Huang2020modeling}, and document-level machine translation \cite{wang2017exploiting}, etc. For these tasks, researchers usually leverage multi-encoder architectures to achieve better performance \cite{zoph2016multi, zhang2018improving, huang2019learning}. To address the data scarcity problem in MSG, some researchers generate pseudo corpora \cite{negri2018escape, nishimura2020multi} to augment the corpus size while others try to make use of pretrained autoencoding models (e.g., BERT \cite{devlin2019bert} and XLM-R \cite{conneau2020unsupervised}) to enhance specific MSG tasks \cite{correia2019simple, lee2020postech, lee2020cross}. Different from these works, we propose a task-agnostic framework to transfer pretrained Seq2Seq models to multi-source sequence generation tasks and demonstrate the generalizability of our framework.

\subsection{Pretraining}
In recent years, self-supervised methods have achieved remarkable success in a wide range of NLP tasks \cite{devlin2019bert, liu2019roberta, conneau2020unsupervised, radford2019language, song2019mass, lewis2020bart, liu2020multilingual}. The architectures of pretrained models can be roughly divided into three categories: autoencoding \cite{devlin2019bert, liu2019roberta, conneau2020unsupervised}, autoregressive \cite{radford2019language}, Seq2Seq \cite{song2019mass, raffel2020exploring, lewis2020bart,liu2020multilingual}. Some researchers propose to use pretrained autoencoding models to improve sequence generation tasks \cite{zhu2019incorporating, guo2020incorporating} and the APE task \cite{correia2019simple}. For pretrained Seq2Seq models, it is convenient to use them to initialize single-source sequence generation models without further modification. Different from these works, we transfer pretrained Seq2Seq models to multi-source sequence generation tasks.

\section{Conclusion}
We propose a novel task-agnostic framework, TRICE, to conduct transfer learning from single-source sequence generation including self-supervised pretraining and supervised generation to multi-source sequence generation. With the help of the proposed gradual finetuning method and the novel MSG model equipped with coarse and fine encoders, our framework outperforms all baselines on three different MSG tasks in four different data scales, which shows the effectiveness and generalizability of our framework.

\section*{Acknowledgments}
This work was supported by the National Key R\&D Program of China (No. 2017YFB0202204), National Natural Science Foundation of China (No.61925601, No. 61772302). We thank all anonymous reviewers for their valuable comments and suggestions on this work.

\bibliographystyle{acl_natbib}
\bibliography{acl2021}

\appendix
\section{Experiment Setup}
\subsection{Model Configurations} \label{app:hyper-para}
We adopted mBART (\emph{mBART-cc25}) \cite{liu2020multilingual} as the pretrained Seq2Seq model. mBART is a Seq2Seq model obtained by multilingual denoising pretraining on a subset of Common Crawl corpus. Following mBART, we set the number of layers of the Coarse-Encoder (i.e., $N_c$) and the number of the Decoder layers (i.e., $N_d$) to 12. Especially, the number of the Fine-Encoder layers (i.e., $N_f$) was set to 1. The model dimension, the filter size, and the number of heads are the same as mBART. We adopted the sentencepiece model provided by mBART for tokenization and adopted the vocabulary of mBART, which contains 250K tokens.

\subsection{Hyper-parameters and Evaluation} 
We used minibatch sizes of 256, 1,024, 4,096, and 16,384 tokens for {\em extremely low-, low-, medium-}, and {\em high-}resource settings, respectively. In each stage of finetuning, we used Adam \cite{kingma2015adam} for optimization and used the learning rate decay policy described by \citet{vaswani2017attention}. We used the development set to tune the hyper-parameters and select the best model. In inference, the beamsize was set to 4 and the length penalty was set to 1.0 , 0.6 and 0 for APE, multi-source translation, and document-level translation, respectively. We used four GeForce RTX 2080Ti GPUs for training. 
We used case-sensitive BLEU ({\em multi-bleu.perl}) and TER\footnote{\url{http://www.cs.umd.edu/˜snover/tercom/}} for automatic post-editing. For multi-source translation and document-level translation, \textsc{SacreBLEU}\footnote{The signature is ``BLEU+case.mixed+numrefs.1+smooth .exp+tok.13a+version.1.4.14"} \cite{post2018call} and METEOR\footnote{\url{https://www.cs.cmu.edu/~alavie/METEOR/}}  was used for evaluation. We used the paired bootstrap resampling \cite{koehn2004statistical} for statistical significance tests.

\subsection{Baselines} \label{app:baselines}
The asterisks (``*") below denote that we report results of these baseline in our implementations in the same hyper-parameter settings as our approach.

\subsubsection*{Automatic Post-Editing}
In the automatic post-editing task, we compare our approach with the following baselines:
\begin{enumerate}
\item \textsc{ForcedAtt} {\cite{berard2017lig}}: a monosource model with a task-specific attention mechanism.
\item \textsc{DualTrans} {\cite{junczys2018ms}}: a dual-source Transformer based model for APE.
\item \textsc{L2Copy} {\cite{huang2019learning}}: a dual-source model enabling cross-source interaction, which focuses on modeling copying mechanism in APE.
\item \textsc{DualBert} {\cite{correia2019simple}}: the first method to use pretrained models to enhance APE, which concatenates multiple sources as a single source and uses two BERT models to initialize the encoder and decoder separately.
\item \textsc{DualBart}* {\cite{correia2019simple}}: adapting \textsc{DualBert} to leverage pretrained Seq2Seq models by concatenating multiple sources as a single source and feeding it to Seq2Seq models.
\end{enumerate}

\subsubsection*{Multi-source Translation}
In the multi-source translation task, we compare our approach with the following baselines:
\begin{enumerate}
\item \textsc{MultiRnn} {\cite{zoph2016multi}}: a multi-source encoder-decoder model based on RNN for machine translation.
\item \textsc{DualTrans}* {\cite{junczys2018ms}}: a dual-source Transformer based model.
\item \textsc{mBart-Trans}* \cite{liu2020multilingual}: a transferring method that directly finetunes the pretrained single-source sequence generation model on the downstream task and takes single-source input during both training and inference.
\item \textsc{DualBart}* {\cite{correia2019simple}}: adapting \textsc{DualBert} to leverage pretrained Seq2Seq models by concatenating multiple sources as a single source and feeding it to the Seq2Seq model.
\end{enumerate}

\subsubsection*{Document-level Translation}
In the document-level translation task, we compare our approach with the following baselines:
\begin{enumerate}
\item SAN {\cite{maruf2019selective}}: a context-aware NMT model with selective attentions.
\item QCN {\cite{yang2019enhancing}}: a context-aware NMT model using a query-guided capsule network.
\item MCN {\cite{zheng2020towards}}: a general-purpose NMT model that is supposed to deal with any-length text.
\item \textsc{mBart-Trans}* \cite{liu2020multilingual}: a transferring method that directly finetunes the pretrained single-source sequence generation model on the downstream task and takes single-source input during both training and inference.
\item \textsc{mBart-DocTrans}* \cite{liu2020multilingual}: a method for document-level translation, which takes $K$ (not more than the number of sentences in a document) source sentences as input and translates $K$ target sentences through a SSG model all at once. For fair comparison, we set $K$ to 2 for both \textsc{mBart-DocTrans} and our approach.
\item \textsc{DualBart}* {\cite{correia2019simple}}: adapting \textsc{DualBert} to leverage pretrained Seq2Seq models by concatenating multiple sources as a single source and feeding it to the Seq2Seq model.
\end{enumerate}

\end{document}